\documentclass[pamm,a4paper,fleqn]{w-art}
\usepackage{times,cite,w-thm}
\usepackage{bm}
\usepackage{amsfonts}
\usepackage[T1]{fontenc}
\usepackage[utf8]{inputenc}
\theoremstyle{plain}

\theoremstyle{definition}

\usepackage{graphicx}

\newcommand*{\tran}{^{\mkern-1.5mu\mathsf{T}}} %
\newcommand{\realNumbers}[1]{\mathbb{R}^{#1}}
\newcommand{\inRealNumbers}[1]{\in \realNumbers{#1}}

\usepackage{pgfplots}
\usepgfplotslibrary{external}
\tikzexternaldisable %
\pgfplotsset{compat=newest}
\usetikzlibrary{patterns, arrows.meta, positioning, spy, calc, shapes.geometric}

\definecolor{clr_1}{RGB}{88,137,176}
\definecolor{clr_2}{RGB}{233,72,73}
\definecolor{clr_3}{RGB}{113,191,110}
\definecolor{clr_4}{RGB}{255,153,51}
\definecolor{clr_5}{RGB}{173,113,181}

\pgfplotsset{
    2dplot/.style={
        axis x line=left,
        axis y line=left,
        legend pos=outer north east,
        legend cell align={left},
        grid=both,
        enlarge x limits,
        enlarge y limits={0.05, upper},
        width=10cm,
        height=6cm,
    },
}

\begin{document}

\TitleLanguage[EN]
\title[]{Towards Using Active Learning Methods for Human-Seat Interactions To Generate Realistic Occupant Motion}

\author{\firstname{Niklas}  \lastname{Fahse}\inst{1,}%
  \footnote{Corresponding author: e-mail \ElectronicMail{niklas.fahse@itm.uni-stuttgart.de}, phone +49\,711\,685\,66560}}
\author{\firstname{Monika} \lastname{Harant}\inst{2}}
\author{\firstname{Marius} \lastname{Obentheuer}\inst{2}}
\author{\firstname{Joachim} \lastname{Linn}\inst{2}}
\author{\firstname{Jörg} \lastname{Fehr}\inst{1}}
\address[\inst{1}]{\CountryCode[DE]Institute of Engineering and Computational Mechanics, University of Stuttgart, Pfaffenwaldring 9, 70569 Stuttgart, Germany}
\address[\inst{2}]{\CountryCode[DE]Fraunhofer Institute for Industrial Mathematics ITWM, Fraunhofer-Platz 1, 67663 Kaiserslautern, Germany}
\AbstractLanguage[EN]
\begin{abstract}
  In the context of developing new vehicle concepts, especially autonomous vehicles with novel seating arrangements and occupant activities, predicting occupant motion can be a tool for ensuring safety and comfort.
  In this study, a data-driven surrogate contact model integrated into an optimal control framework to predict human occupant behavior during driving maneuvers is presented.
  High-fidelity finite element simulations are utilized to generate a dataset of interaction forces and moments for various human body configurations and velocities.
  To automate the generation of training data, an active learning approach is introduced, which iteratively queries the high-fidelity finite element simulation for an additional dataset.
  The feasibility and effectiveness of the proposed method are demonstrated through a case study of a head interaction with an automotive headrest, showing promising results in accurately replicating contact forces and moments while reducing manual effort.
\end{abstract}

\maketitle

\section{Introduction}
The development of new vehicle concepts, especially in the context of autonomous vehicles where novel seating arrangements and occupant activities are possible \cite{SunCaoTang21}, is a complex task that requires the evaluation of occupant safety and comfort.
To reduce costs, virtual human body models (HBMs) are used in all development phases to predict the passive and active behavior of the occupants in different scenarios.
For evaluating ergonomics and comfort, e.g., RAMSIS \cite{BubbEtAl06} is used in posture prediction.
Finite element (FE) simulations are, e.g., now part of the pedestrian protection evaluation in the Euro NCAP (New Car Assessment Programme) \cite{KlugEllway21} assessment.
In the EMMA4Drive project \cite{ObentheuerEtAl23}, the authors aim to predict human occupant behavior during driving maneuvers by extending the muscle-activated multibody model EMMA (Ergo-dynamic Moving Manikin) \cite{RollerEtAl20} to ultimately evaluate the safety and comfort of the occupants in autonomous vehicles.
Using optimal control, the multibody system (MBS) model is actuated to perform tasks such as reaching a certain position or executing a specific movement.
Since occupant motion is considered, the human body is in continuous contact with the vehicle seat.
This complex and widespread interaction must be accounted for in the optimal control problem (OCP) to accurately predict human occupant behavior.
This contribution describes a data-driven surrogate model approach to include the occupant-seat interaction in the OCP.
High-fidelity FE simulations are employed to create the training dataset of contact forces for various human body configurations and velocities.
To overcome the high manual effort of generating training data, an active learning approach is introduced to automate the generation of meaningful relative trajectories of the human body and the seat.
These trajectories are then automatically evaluated in the FE simulation to generate training data for the surrogate model.

\section{Methods\label{sec:methods}}
\subsection{Human Behavior Prediction with an Optimally Controlled Biomechanical Multibody Model}
In this work, human behavior is predicted using an optimally controlled biomechanical MBS model resembling the human body.
This procedure is described in detail in \cite{RollerEtAl17}.
Considering a kinematic tree with $N$ rigid bodies connected by $N$ joints, a configuration can be described by the vector of the joint coordinates~$\bm{q} \inRealNumbers{n}$ with the total number of degrees of freedom (DOF)~$n$ calculated as the sum of the DOF~$n_i$ of each joint.
When subjected to external forces~$\bm{f}^{\text{ext}} \inRealNumbers{n}$ and joint actuations~$\bm{\tau} \inRealNumbers{n}$, the equations of motion are given by
\begin{equation}
  \begin{aligned}
    \label{eq:equation_of_motion}
    \bm{M}(\bm{q})\ddot{\bm{q}} + \bm{C}(\bm{q},\dot{\bm{q}}) - \bm{G}(\bm{q})\tran \bm{\lambda} & = \bm{\tau} + \bm{f}^{\text{ext}}, \\
    \bm{g}(\bm{q})                                                                               & = \bm{0}
  \end{aligned}
\end{equation}
where $\bm{M}(\bm{q}) \inRealNumbers{n \times n}$ is the joint-space mass matrix and $\bm{C}(\bm{q},\dot{\bm{q}}) \inRealNumbers{n}$ is the Coriolis and centrifugal forces term.
The partial derivative~$\bm{G}(\bm{q}) = \frac{\partial \bm{g}(\bm{q})}{\partial \bm{q}} \in \mathbb{R}^{m \times n}$ of the constraint function~$\bm{g}(\bm{q}) \in \mathbb{R}^m$ appears if kinematic loops occur and is associated with the Lagrangian multipliers \(\bm{\lambda} \in \mathbb{R}^m\).
Hereby, $m$ is the number of constraints due to kinematic loops.

The overall goal is to predict human behavior, i.e., the evolution of joint actuations~$\bm{\tau}$ in a time frame $t_0$ to $t_\textrm{F}$, given a certain task.
Thus, from considering an instance of time in~\eqref{eq:equation_of_motion}, $\bm{q}$, $\bm{\tau}$, and $\bm{\lambda}$ become the functions $\bm{q}: [t_0, t_\textrm{F}] \rightarrow \realNumbers{n}$, $\bm{\tau}: [t_0, t_\textrm{F}] \rightarrow \realNumbers{n}$, and $\bm{\lambda}: [t_0, t_\textrm{F}] \rightarrow \realNumbers{m}$ on the time interval~$[t_0, t_\textrm{F}]$.
Using optimal control, the joint actuations~$\bm{\tau}$ are determined by minimizing a cost function~$J$. The OCP used in this study may be formulated as
\begin{align}
  \min_{\bm{\tau}} J                                                                                                                                                              & = E(\bm{q}|_{t_0}, \dot{\bm{q}}|_{t_0}, t_0, \bm{q}|_{t_{\textrm{F}}}, \dot{\bm{q}}|_{t_{\textrm{F}}}, t_{\textrm{F}}) + \int_{t_0}^{t_\textrm{F}} F(\bm{q}, \dot{\bm{q}}, \bm{\tau}, \bm{\lambda}, t) \, \mathrm{d}t, \label{eq:optimal_control_problem_cost}                                    \\
  \text{s.t.} \quad \bm{M}(\bm{q})\ddot{\bm{q}} + \bm{C}(\bm{q},\dot{\bm{q}}) - \bm{G}(\bm{q})\tran \bm{\lambda}                                                                  & = \bm{\tau} + \bm{f}^{\text{ext}},                                                                                                                                                                       \label{eq:optimal_control_problem_eom}                                                   \\
  \bm{g}(\bm{q})                                                                                                                                                                  & = \bm{0},                                                                                                                                                                                                                                      \label{eq:optimal_control_problem_eom_constraints} \\
  \bm{c}^- \leq                                                                                                              \bm{c}(\bm{q},\dot{\bm{q}}, \bm{\tau}, \bm{\lambda}) & \leq \bm{c}^+. \label{eq:optimal_control_problem_constraints}
\end{align}
The cost function~$J$ introduced in~\eqref{eq:optimal_control_problem_cost} is composed of an endpoint cost~$E$ and a running cost~$F$.
In this study, a running cost~$F$ is used as a weighted sum of cost functions to prioritize minimal control and minimal kinetic energy~$T$, i.e.,
\begin{equation}
  \label{eq:running_cost}
  F = w_{\mathrm{\tau}} \, \bm{\tau}\tran \bm{\tau} + w_{\textrm{T}} \, T
\end{equation}
with the weights~$w_{\textrm{u}}$, and~$w_{\textrm{T}}$.
The minimization problem is subject to the constrained equations of motion~\eqref{eq:optimal_control_problem_eom}-\eqref{eq:optimal_control_problem_eom_constraints} as well as additional equality and inequality constraints~\eqref{eq:optimal_control_problem_constraints}.

To include changing objective functions and constraints, the OCP is split into multiple phases.
The overall time interval~$I = [t_0, t_\textrm{F}]$ is divided into $M \in \mathbb{N}$ sub-intervals
\begin{equation}
  I_m = [t_{0,m}, t_{\textrm{F},m}] \in I
\end{equation}
with $t_{\textrm{F},m} = t_{0,m+1} \textrm{ for } m \in \{1, \ldots, M-1\}$.
This introduces additional constraints to the OCP to ensure, e.g., continuity of the state variables at the phase boundaries.
The OCP as described in~\eqref{eq:optimal_control_problem_cost}-\eqref{eq:optimal_control_problem_constraints} is formulated on continuous functions and thus forms an infinite-dimensional optimization problem.
To solve the OCP, the DMOCC (discrete mechanics and optimal control for constrained systems) method from \cite{LeyendeckerEtAl10} is used to discretize the problem.
The resulting finite-dimensional optimization problem is solved using the solver IPOPT \cite{WachterBiegler06}.
Roller et al. \cite{RollerEtAl17} use constraints appearing as additional equations in~\eqref{eq:optimal_control_problem_eom_constraints} to model rigid body contacts.
Using the concept of phases, the opening and closing of rigid body contacts are addressed by activating constraints only in specific phases. Moreover, \cite{RollerEtAl17} extends this method to allow arbitrary contact points and sticking contacts.

\subsection{Surrogate Model of Occupant-Seat Interaction\label{sec:surrogate_model}}
When considering the widespread and complex contact of the human body with the seat, the methods for rigid body contacts described in Roller et al. \cite{RollerEtAl17} are not applicable.
This work introduces a surrogate model approach to include the occupant-seat interaction in the OCP.
Therefore, equation~\eqref{eq:optimal_control_problem_eom} is extended by the contact forces~$\bm{f}^{\text{cm}}(\bm{q}, \dot{\bm{q}}) \inRealNumbers{n}$ acting on the human body due to the contact with the seat, resulting in the equations of motion
\begin{equation}
  \label{eq:equation_of_motion_contact}
  \bm{M}(\bm{q})\ddot{\bm{q}} + \bm{C}(\bm{q},\dot{\bm{q}}) - \bm{G}(\bm{q})\tran \bm{\lambda} = \bm{\tau} + \bm{f}^{\text{ext}} + \bm{f}^{\text{cm}}(\bm{q}, \dot{\bm{q}}).
\end{equation}
To keep the benefits of gradient-based optimization, the contact surrogate model is required to be differentiable with respect to the state variables~$\bm{q}$ and $\dot{\bm{q}}$ and be computationally efficient.
In this work, a data-driven approach is used to create the contact surrogate model.
Therefore, in an offline phase, high-fidelity FE simulations are conducted to generate a dataset of contact forces~$\bm{f}^{\text{FE}}$ for various human body configurations~$\bm{q}^\text{FE}$ and velocities~$\dot{\bm{q}}^\text{FE}$.
Hereby, $\Box^\text{FE}$ denotes the values obtained from the FE simulation.
The mapping $(\bm{q}^\text{FE}, \dot{\bm{q}}^\text{FE}) \mapsto \bm{f}^{\text{FE}}$ is then approximated using a machine learning (ML) model, i.e.,
\begin{equation}
  \label{eq:surrogate_model}
  \bm{f}^{\text{ML}}(\bm{q}^\text{FE}, \dot{\bm{q}}^\text{FE}) \approx \bm{f}^{\text{FE}}
\end{equation}
trained on the dataset generated in the offline phase.
Assuming that the FE simulation and the MBS model describe the same physical system, the surrogate model~$\bm{f}^{\text{ML}}$ is used for the contact forces~$\bm{f}^{\text{cm}}$ in the OCP, i.e., $\bm{f}^{\text{cm}}(\bm{q}, \dot{\bm{q}}) = \bm{f}^{\text{ML}}(\bm{q}, \dot{\bm{q}})$.
With this procedure, the interaction of the human body model with the seat in the FE simulation is transferred to the optimally controlled MBS model --~this is the online phase~-- by using the surrogate model as an interface.
Of course, the FE model and the MBS model have to be matched in terms of the anthropometry and coordinate systems to describe the kinematics.
Here, coordinate systems based on the PIPER framework \cite{PIPER24} are used in both models and the MBS model is adjusted to the anthropometry of the FE model.
The human body is then split into body parts with respective contact regions and coordinate systems describing the kinematics and a contact surrogate model is trained for each pair of contact opponents.
By not using one model for the whole body but multiple models for each contact pair, the complexity of the ML model is reduced and the training data can be generated more efficiently by considering only one contact pair at a time.

A central challenge in this approach is the generation of the training data.
Without any knowledge of the online phase, the training data has to cover the whole possible range of occupant-seat interactions.
Considering the dimensionality of the relative kinematics and the high computational cost of FE simulations, this is not feasible.
Even with some prior knowledge of possible relative configurations, an expert would have to generate trajectories manually to set up the transient simulations of the offline phase.
To overcome these challenges, an active learning approach is used to generate the training data and train the ML model simultaneously.

\subsection{An Active Learning Approach for Simultaneous Training Data Generation and ML Model Training\label{sec:active_learning}}
\begin{figure}[ht]
  \centering
  \tikzsetnextfilename{activelearningloop}
  \begin{tikzpicture}[
    node distance=2cm and 2cm,
    task/.style={
            draw,
            rectangle,
            rounded corners,
            minimum width=3.3cm,
            minimum height=2.4cm,
            inner sep=0mm,
            align=center,
            label={[anchor=west,fill=white,xshift=2mm] north west:#1}
        },
    arrow/.style={-{>[length=1.5mm]}, shorten >=2mm, shorten <=2mm},
    twosidedarrow/.style={{<[length=1.5mm]}-{>[length=1.5mm]}, shorten >=2mm, shorten <=2mm},
    ]
    \pgfdeclareimage[width=2cm]{dataset}{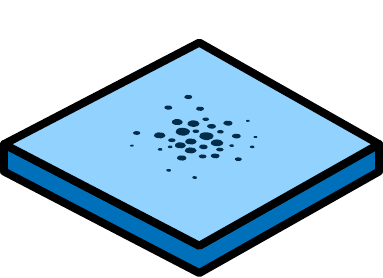};
    \pgfdeclareimage[width=2cm]{prelimdataset}{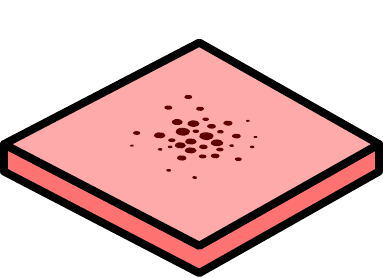};
    \pgfdeclareimage[width=2cm]{newdataset}{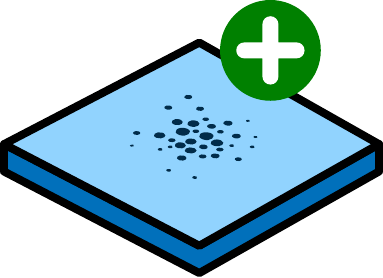};
    \pgfdeclareimage[width=2cm]{femesh}{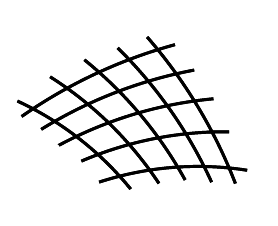};

    \node[task={Data pool}, inner sep=-10mm] (Data Pool) {
        \begin{tikzpicture} [
                node distance=0cm and 0cm
            ]
            \node[inner sep=0pt] (dataset) at (0, 0) {\pgfuseimage{prelimdataset}};
            \node[inner sep=0pt] (dataset) at (0, 0.3)  {\pgfuseimage{dataset}};
            \node[inner sep=0pt] (dataset) at (0, 0.6)  {\pgfuseimage{dataset}};
        \end{tikzpicture}
    };
    \node[task={Surrogate model}, right=of Data Pool] (Surrogate Model) {
        \begin{tikzpicture} [scale=0.4]
            \filldraw[fill=blue!60!black, draw=blue!60!black]
            (-1.5, 0.5) circle (0.3)
            (-1.5, 1.5) circle (0.3)
            (0, 0) circle (0.3)
            (0, 1) circle (0.3)
            (0, 2) circle (0.3)
            (1.5, 1) circle (0.3);
            \draw[draw=blue!60!black, very thick] (-1.5, 0.5) -- (0, 0);
            \draw[draw=blue!60!black, thick] (-1.5, 1.5) -- (0, 0);
            \draw[draw=blue!60!black, thick] (-1.5, 1.5) -- (0, 1);
            \draw[draw=blue!60!black, very thick] (-1.5, 0.5) -- (0, 2);
            \draw[draw=blue!60!black, thick] (-1.5, 1.5) -- (0, 2);
            \draw[draw=blue!60!black, thick] (-1.5, 0.5) -- (0, 1);
            \draw[draw=blue!60!black, very thick ] (0, 0) -- (1.5, 1);
            \draw[draw=blue!60!black, thick] (0, 1) -- (1.5, 1);
            \draw[draw=blue!60!black, thick] (0, 2) -- (1.5, 1);
        \end{tikzpicture}
    };
    \node[task={MBS simulation}, below=of Surrogate Model] (MBS Simulation) {
        \begin{tikzpicture}
            \draw[thick, rotate=100] (0, 0) ellipse (0.5cm and 0.2cm);

            \begin{scope}[shift={(-0.16,-0.9)}, sharp corners]
                \draw[line width=1pt] (0,0) -- (0.5,0)
                -- (0.25,0.4) -- cycle;

                \foreach \x in {0,...,5} {
                        \draw[line width=.75pt] (\x/5*0.5,0) -- (\x/5*0.5+0.1,-0.1);
                    }

                \filldraw[line width=1pt,  fill=white] (0.25,0.4) circle (0.1);
            \end{scope}
            \draw[thick, rotate=-20] (0.8cm, 0.1cm) ellipse (0.5cm and 0.2cm);
            \filldraw[line width=1pt,  fill=white] (0.25,0) circle (0.1);
        \end{tikzpicture}
    };
    \node[task={FE simulation}, left=of MBS Simulation, inner sep=-10mm] (FE Simulation) {
        \begin{tikzpicture}
            \node[inner sep=0pt] (dataset) {\pgfuseimage{newdataset}};
            \node[inner sep=0pt, anchor=center, xshift=-0.8cm, yshift=0.3cm] at (dataset.center) {\pgfuseimage{femesh}};
        \end{tikzpicture}
    };
    \node[left=of Data Pool,align=center, xshift=0.5cm] (Start) {
        \begin{tikzpicture} [
                node distance=0cm and 0cm
            ]
            \node[inner sep=0pt] (dataset) {\pgfuseimage{prelimdataset}};
            \node[below=of dataset] {Preliminary data};
        \end{tikzpicture}
    };
    \node[right=of MBS Simulation,align=center, xshift=-0.5cm] (Task) {
        \begin{tikzpicture} [
                node distance=0cm and 0cm
            ]
            \node (sheet) {
                \begin{tikzpicture}
                    \def\height{1.4142cm}
                    \def\width{1cm}
                    \def\cornerwidth{0.3cm}
                    \foreach \i in {0.2,0.4,...,1.2} {
                            \draw (-\width + 0.1cm, 0 cm) ++(0,\i cm - 0.05cm) rectangle ++(0.1cm,0.1cm);
                            \draw (-\width + 0.3cm, 0 cm) ++(0,\i cm - 0.05cm) -- ++(0.5cm,0);
                        }
                    \filldraw[fill=white, thick] (0, \height - \cornerwidth) -- ++(-\cornerwidth,0) -- ++(0,\cornerwidth);
                    \draw[thick] (0,0)
                    --  ++(-\width,0)
                    --  ++(0,\height)
                    --  ++(\width-\cornerwidth,0)
                    --  ++(\cornerwidth,-\cornerwidth)
                    -- cycle;
                \end{tikzpicture}
            };
            \node[below=of sheet] {Task definition};
        \end{tikzpicture}
    };

    \draw[arrow] (Data Pool) -- (Surrogate Model) node[midway, above, align=center] {Train\\model};
    \draw[arrow] (Surrogate Model) -- (MBS Simulation) node[midway, left, align=center] {Use\\model};
    \draw[arrow] (MBS Simulation) -- (FE Simulation) node[midway, above, align=center] {Use\\trajectory};
    \draw[arrow] (FE Simulation) -- (Data Pool) node[midway, left, align=center] {Append to\\data pool};
    \draw[arrow] (Start) -- (Data Pool);
    \draw[arrow] (Task) -- (MBS Simulation);

    \draw[twosidedarrow, dashed, black!80] (MBS Simulation) to[bend left] coordinate[pos=.5] (M) (FE Simulation);
    \node[align=center, yshift=-0.5cm, black!80] at (M) {Evaluate\\error};
\end{tikzpicture}
  \caption{Active learning approach for generating training data and training the ML model.}
  \label{fig:activelearningloop}
\end{figure}
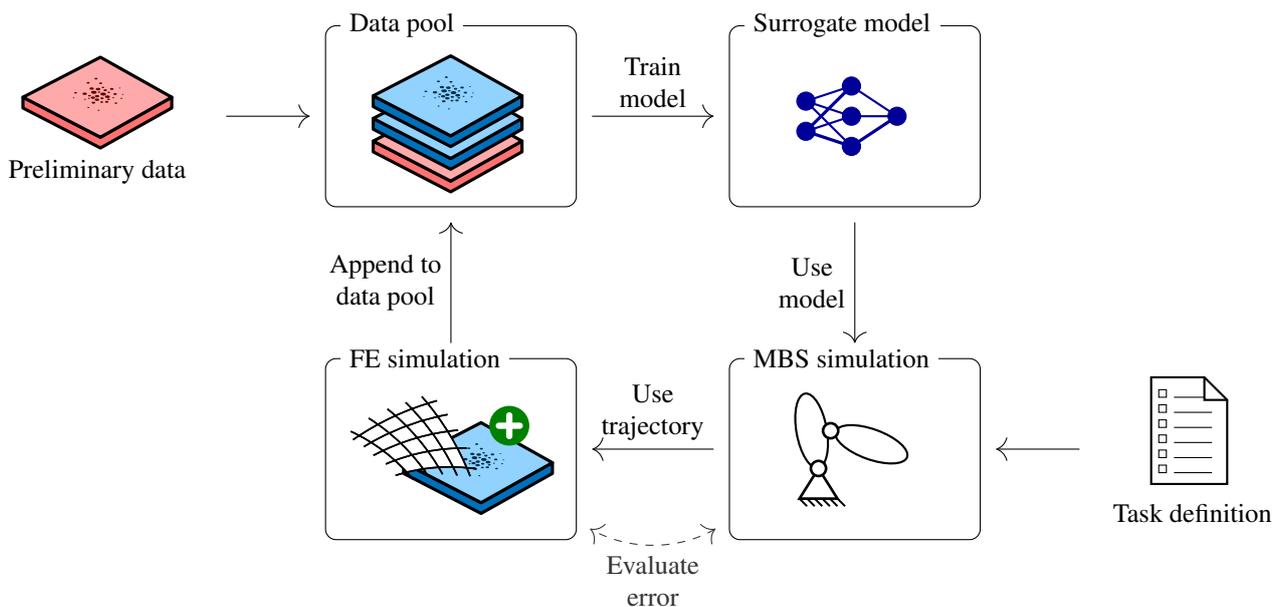

The key idea behind active learning is that the learning algorithm chooses the data from which it learns \cite{Settles09}.
Therefore, the learning algorithm is allowed to ask \textit{queries} to an \textit{oracle} which is considered the source of the ground truth.
There are different strategies to choose the queries, e.g., uncertainty sampling, query-by-committee, or expected model change, see \cite{Settles09} for an overview. 
These strategies all have in common that they are motivated by finding queries that are most informative to the learning algorithm to reduce the high cost of labeling data, e.g., in medical imaging \cite{BuddRobinsonKainz21}.
In contrast, active learning in this work is used to automate the generation of queries in the form of meaningfully relative trajectories of the human body and seat with the help of the intermediate ML model.
This is due to the fact, that not the labeling of the data, i.e., calculating the evolution of contact forces for a trajectory, but finding realistic contact trajectories requires manual effort and thus is costly.
The proposed active learning approach leads to an iterative process of training the ML model and generating new data, see Figure~\ref{fig:activelearningloop}.
Starting from a preliminary dataset a surrogate model is trained.
This intermediate model is then used to approximate the occupant-seat interaction in the former online phase.
Hereby, a task definition specifies the cost function and constraints for the optimally controlled MBS model.
The resulting trajectories are then replayed in the FE simulation to generate new data which is appended to the data pool.
This closes the loop and the surrogate model is retrained using the extended dataset.
The process is repeated until a stopping criterion is met.
With Figure~\ref{fig:activelearningloop} it is directly visible that the iterative process relies on preliminary data to be started.
This preliminary data can be generated using a very simple heuristic, e.g. some penalty-based approach, or by using a small set of manually generated trajectories in the FE simulation.
The active learning approach then iteratively adds datasets originating from the FE simulation -- which is considered the ground truth -- to the data pool.

\section{Modeling and Results}
As a proof of concept, the active learning approach is applied to a simple head-neck model with only one degree of freedom which interacts with an automotive headrest.
The task under investigation is to move the head from a fixed initial position to the headrest.
This is a task frequently performed in everyone's daily life, not only as a passenger in a vehicle but could also be replicated while reading this contribution.
It seems trivial at first glance, but even with this small movement, there is a preferred way of doing it in terms of comfort and effort:
How hard does the head hit the headrest and how large are the head accelerations (how much comfort)?
How much of the potential energy of the head is taken out of the system by the joint actuation in comparison to dissipation in the headrest cushion (how much effort)?
These questions reveal a conflict of objectives that is -- in this study -- resolved by formulating a cost function in an OCP. 
Even though the selected task is interesting in itself, the main goal is to show the feasibility of the active learning approach in this proof of concept and therefore to keep the computation times -- especially for the FE simulations -- low.
The following sections describe the MBS model, the FE model, and the ML model used in this proof of concept. In Section~\ref{sec:active_learning_loop}, the results of the active learning loop are presented.

\subsection{Multibody Model}
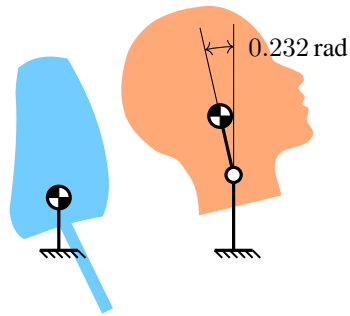
\begin{figure}[ht]
  \centering
  \tikzsetnextfilename{multibodymodel}
  \begin{tikzpicture}[
    node distance=2cm and 2cm,
    task/.style={
            draw,
            rectangle,
            rounded corners,
            minimum width=3cm,
            align=center,
            label={[anchor=west,fill=white,xshift=2mm] north west:#1}
        },
    arrow/.style={-{>[length=1.5mm]}},
    ]

    \pgfdeclareimage[width=4cm]{mbs}{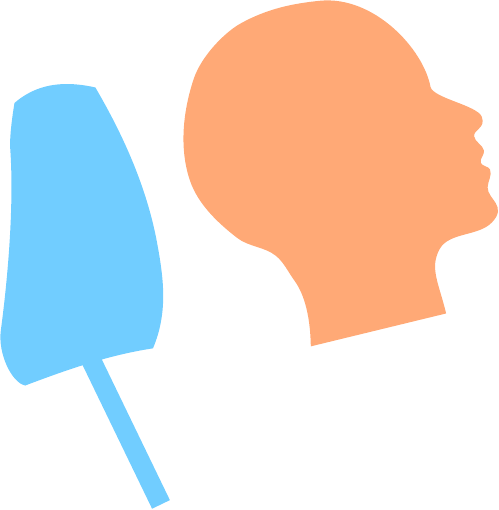};

    \node at (-0.7cm, 1.2cm) {\pgfuseimage{mbs}};

    \begin{scope}[shift={(0,0)}]
        \draw[very thick] (0,0) -- (0.5,0);
        \foreach \x in {0,...,5} {
                \draw[line width=.75pt] (\x/5*0.5,0) -- (\x/5*0.5+0.1,-0.1);
            }
    \end{scope}

    \draw[very thick] (0.25,0) -- (0.25, 1cm);

    \draw[very thick] (0.25, 1cm) -- ++({-0.8*0.221156329}, {0.8*0.9752383699}); %

    \draw[] (0.25, 1cm) -- (0.25, 1cm + 2cm);
    \draw[] (0.25, 1cm) -- ++({-2*0.221156329}, {2*0.9752383699});
    \draw[<->] (0.25, 1cm + 1.7cm) arc (90:90+12.77695:1.7cm);
    \node at (0.25 + 1.7*0.5, 1cm + 1.7cm) {$0.232\,\textrm{rad}$};

    \filldraw[very thick,  fill=white] (0.25, 1cm) circle (0.1);

    \begin{scope}[shift={({0.25-0.8*0.221156329}, {1 + 0.8*0.9752383699})}]
        \filldraw[very thick,  fill=white] (0,0) circle (0.15);
        \filldraw[fill=black] (0,0) -- (0.15,0) arc (0:-90:0.15) -- cycle;
        \filldraw[fill=black] (0,0) -- (-0.15,0) arc (180:90:0.15) -- cycle;
    \end{scope}

    \begin{scope}[shift={(-2.3,0)}]
        \draw[very thick] (0,0) -- (0.5,0);
        \foreach \x in {0,...,5} {
                \draw[line width=.75pt] (\x/5*0.5,0) -- (\x/5*0.5+0.1,-0.1);
            }
    \end{scope}

    \draw[very thick] (-2.05,0) -- (-2.05, 0.7cm);

    \begin{scope}[shift={(-2.05, 0.7cm)}]
        \filldraw[very thick,  fill=white] (0,0) circle (0.15);
        \filldraw[fill=black] (0,0) -- (0.15,0) arc (0:-90:0.15) -- cycle;
        \filldraw[fill=black] (0,0) -- (-0.15,0) arc (180:90:0.15) -- cycle;
    \end{scope}

\end{tikzpicture}
  \caption{Kinematic structure of the MBS model with one degree of freedom.}
  \label{fig:multibodymodel}
\end{figure}

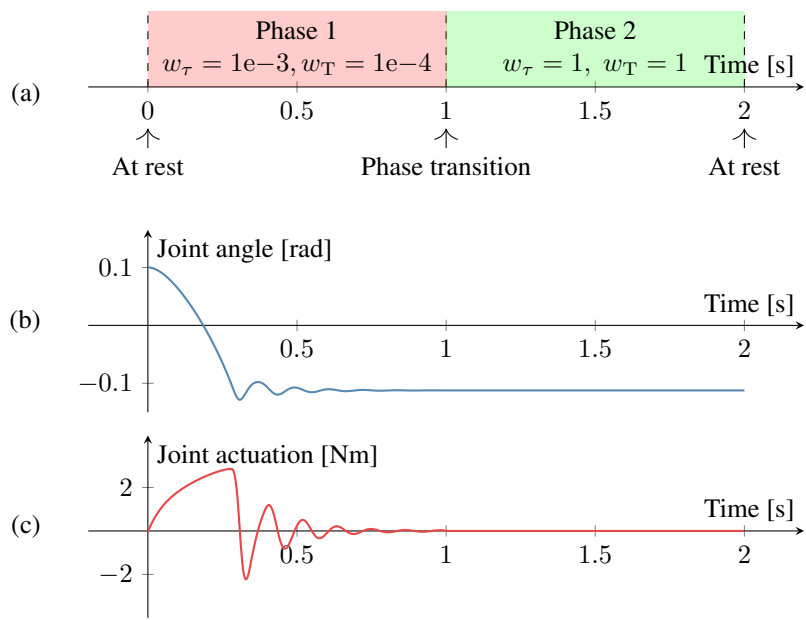
\begin{figure}[ht]
  \centering
  \tikzsetnextfilename{mb_phases}
  \begin{tikzpicture}[
    scale=1,
    arrow/.style={-{>[length=1.5mm]}},
    ]
    \begin{axis}
        [
            xlabel={Time [s]},
            xmin=0, xmax=2,
            ymin=-1.3, ymax=1,
            axis x line=middle,
            enlarge x limits,
            enlarge y limits={0.05, upper},
            width=11cm,
            height=4cm,
            xtick={0,0.5,1,1.5,2},
            hide y axis,
            name=phases,
        ]
        \addplot[
            color=blue,
            thick,
            domain=0:2,
            transparent,
        ]
        {sin(deg(2*pi*x))};
        \node[anchor=south, align=center] (trans) at (axis cs:1, -1.3) {Phase transition};
        \draw[arrow] (trans.north) -- (axis cs:1, -0.55);
        \draw[dashed] (axis cs:1, 0) -- (axis cs:1, 1);
        \fill[red, opacity=0.2] (axis cs:0, 0) rectangle (axis cs:1, 1);
        \fill[green, opacity=0.2] (axis cs:1, 0) rectangle (axis cs:2, 1);
        \node[anchor=north, align=center] at (axis cs:0.5, 1) {Phase 1\\$w_{\mathrm{\tau}} = 1\mathrm{e}{-3},w_{\mathrm{T}} = 1\mathrm{e}{-4}$};
            \node[anchor=north, align=center] at (axis cs:1.5, 1) {Phase 2\\$w_{\mathrm{\tau}} = 1,\;w_{\mathrm{T}} = 1$};
        \node[anchor=south, align=center] (rest1) at (axis cs:0, -1.3) {At rest};
        \draw[arrow] (rest1.north) -- (axis cs:0, -0.55);
        \draw[dashed] (axis cs:0, 0) -- (axis cs:0, 1);
        \node[anchor=south, align=center] (rest2) at (axis cs:2, -1.3) {At rest};
        \draw[arrow] (rest2.north) -- (axis cs:2, -0.55);
        \draw[dashed] (axis cs:2, 0) -- (axis cs:2, 1);
    \end{axis}
    \node[
        anchor=east,
        xshift=-0.5cm,
    ] at (phases.west) {(a)};
    \begin{axis}
        [
            at={(phases.below south west)},yshift=-0.6cm,
            anchor=north west,
            xlabel={Time [s]},
            ylabel={Joint angle [rad]},
            xmin=0, xmax=2,
            xtick={0,0.5,1,1.5,2},
            ymin=-0.15, ymax=0.15,
            axis lines=left,
            axis lines=middle,
            enlarge x limits,
            enlarge y limits={0.05, upper},
            width=11cm,
            height=4cm,
            name=coordinateplot,
        ]
        \addplot[
            color=clr_1,
            thick,
        ]
        table[x=time, y=angle, col sep=comma] {img/mbs_sim.dat};
        \node[above left] at (rel axis cs:0,1) {(b)};
    \end{axis}
    \node[
        anchor=east,
        xshift=-0.5cm,
    ] at (coordinateplot.west) {(b)};
    \begin{axis}
        [
            at={(coordinateplot.below south west)},yshift=-0.3cm,
            anchor=north west,
            xlabel={Time [s]},
            xtick={0,0.5,1,1.5,2},
            ylabel={Joint actuation [Nm]},
            xmin=0, xmax=2,
            ymin=-4, ymax=4,
            ytick={-2,0,2},
            axis lines=left,
            axis lines=middle,
            enlarge x limits,
            enlarge y limits={0.05, upper},
            width=11cm,
            height=4cm,
            name=actuationplot,
        ]
        \addplot[
            color=clr_2,
            thick,
        ]
        table[x=time, y=actuation, col sep=comma] {img/mbs_sim.dat};
        \node[above left] at (rel axis cs:0,1) {(c)};
    \end{axis}
    \node[
        anchor=east,
        xshift=-0.5cm,
    ] at (actuationplot.west) {(c)};
\end{tikzpicture}
  \caption{Multibody simulation of the head movement toward the headrest. (a) Phases of the simulation with varying cost functions. (b) Joint angle and (c) joint actuation of an exemplary solution.}
  \label{fig:mb_phases}
\end{figure}

The MBS model used for this proof of concept is a simple head-neck model with one degree of freedom, see Figure~\ref{fig:multibodymodel}.
The headrest is fixed in space and the neck movement is modeled as a revolute joint positioned between the C7 and T1 vertebrae.
The range of motion of the joint spans $0.4\,\textrm{rad}$ and excludes the position where the head center of gravity (COG) is directly above the joint.
This way, the only possibility to reach a position at rest without any actuation is to lay the head on the headrest.
The model is actuated using a torque actuator at the joint which is limited to $\pm30\,\textrm{Nm}$.
The simulation spans $2\,\textrm{s}$ and is separated into two phases, see Figure~\ref{fig:mb_phases}~(a).
The first phase starts at $t=0\,\textrm{s}$ with the head in a resting initial position where the head COG is tilted $0.232\,\textrm{rad}$ toward the headrest, see Figure~\ref{fig:multibodymodel}.
In the first phase, the cost function has a small weight on the joint actuation and kinetic energy, i.e., $w_{\mathrm{\tau}} = 1\mathrm{e}{-3}$ and $w_{\mathrm{T}} = 1\mathrm{e}{-4}$, to allow the head to move toward the headrest.
At the transition to the second phase at $t=1\,\textrm{s}$, the cost function is changed to higher weights to guide the solution toward a sustained position at rest, i.e., $w_{\mathrm{\tau}} = 1$ and $w_{\mathrm{T}} = 1$.
By the end of the second phase at $t=2\,\textrm{s}$, the head is constrained to be at rest again.
With this two-phase setup, the head reaches the headrest and comes to rest within the first phase.
The second phase then ensures that the head remains at rest without any actuation.
When instead using only terminal constraints, the solution exploits a dynamic equilibrium to meet the constraints.
In Figures~\ref{fig:mb_phases}~(b)~and~(c), the joint angle and actuation of an exemplary solution to the OCP are shown for the first and second phases.
The solution is found using the method described in Section~\ref{sec:methods} and shows a movement of the head toward the headrest and a subsequent resting with no actuation in the second phase.
Especially before the first contact, it is visible that the actuation serves to counteract the gravitational force and to reduce the kinetic energy, i.e. the velocity of the head.

\subsection{Finite Element Model}
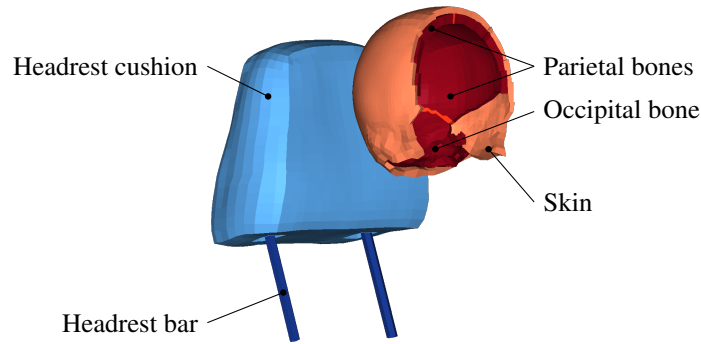
\begin{figure}[ht]
  \centering
  \tikzsetnextfilename{FEsim}
  \begin{tikzpicture}[
    ]
    \pgfdeclareimage[width=5cm]{fesim}{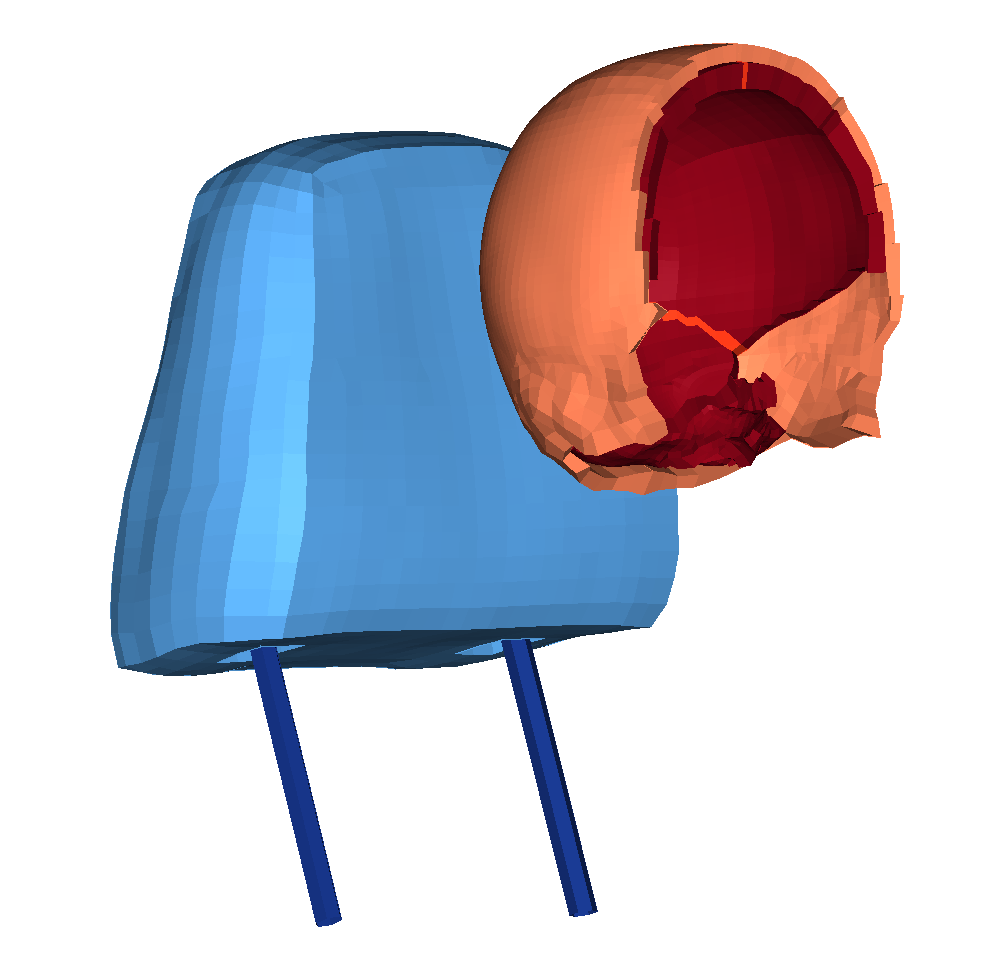};

    \node (dataset) at (0, 0) {\pgfuseimage{fesim}};

    \node[align=right, anchor=east] (headrestbar) at (-2cm, -2cm) {Headrest bar};
    \draw (headrestbar.east) -- (-1cm, -1.6cm);
    \filldraw (-1cm, -1.6cm) circle (1pt);

    \node[align=right, anchor=east] (headrest) at (-2cm, 1.4cm) {Headrest cushion};
    \draw (headrest.east) -- (-1.2cm, 1cm);
    \filldraw (-1.2cm, 1cm) circle (1pt);

    \node[align=left, anchor=west] (skull1) at (2.3cm, 1.4cm) {Parietal bones};
    \draw (skull1.west) -- (1.2cm, 1cm);
    \filldraw (1.2cm, 1cm) circle (1pt);
    \draw (skull1.west) -- (0.95cm, 1.9cm);
    \filldraw (0.95cm, 1.9cm) circle (1pt);

    \node[align=left, anchor=west] (skull2) at (2.3cm, 0.8cm) {Occipital bone};
    \draw (skull2.west) -- (1cm, 0.3cm);
    \filldraw (1cm, 0.3cm) circle (1pt);

    \node[align=left, anchor=west] (skin) at (2.3cm, -0.4cm) {Skin};
    \draw (skin.west) -- (1.7cm, 0.3cm);
    \filldraw (1.7cm, 0.3cm) circle (1pt);

\end{tikzpicture}
  \caption{Reduced FE model of the head and the headrest. The headrest consists of a cushion and a headrest bar. The head is reduced to the parietal and occipital bones and the covering skin.}
  \label{fig:FEsim}
\end{figure}

As described in Section~\ref{sec:surrogate_model}, the goal of the FE simulations is to calculate the nodal interaction forces when the head comes into contact with the headrest.
The FE model used in this work is based on the THUMS (Total Human Model for Safety) model in the male $50^\textrm{th}$ percentile occupant v4.1 configuration \cite{IwamotoNakahiraKimpara15} and a model of a 2012 Toyota Camry passenger sedan \cite{MarzouguiEtAl14}.
These models are both open-accessible and are available for the LS-DYNA \cite{Hallquist06} simulation software.
The THUMS model is reduced to the parietal and occipital bones of the head and the corresponding covering skin, see Figure~\ref{fig:FEsim}.
The car model is also greatly simplified by using the front passenger headrest which consists of a cushion part and a headrest bar.
The cushion is modeled with \texttt{*MAT\_LOW\_DENSITY\_FOAM} and features a highly nonlinear stress-strain curve.
The headrest bar and the bones are rigidified and can be moved independently in space using \texttt{*BOUNDARY\_PRESCRIBED\_MOTION\_RIGID} to simulate arbitrary contact situations.
This further reduces the computational cost of the FE simulation, and, because the loads are expected to be very low compared to the stiffness of the bones, does not affect the result.
The removed parts of the head model and the car model don't distort the result because the rigidified headrest bar and skull are moved by the boundary conditions.
Therefore, the dynamics of other parts besides the head skin and the headrest cushion do not affect the interaction forces, which are the only quantities of interest.

\begin{figure}[ht]
  \centering
  \tikzsetnextfilename{FEmovement}
  \pgfplotsset{
    myplot/.style={
            xlabel={Time [s]},
            xmin=0, xmax=1,
            xtick={0, 0.29,0.5, 1},
            axis lines=left,
            axis lines=middle,
            enlarge x limits,
            enlarge y limits={0.05, upper},
            width=8cm,
            height=3.2cm,
        },
}

\begin{tikzpicture}[
    ]
    \pgfdeclareimage[width=4.5cm]{FEmovement}{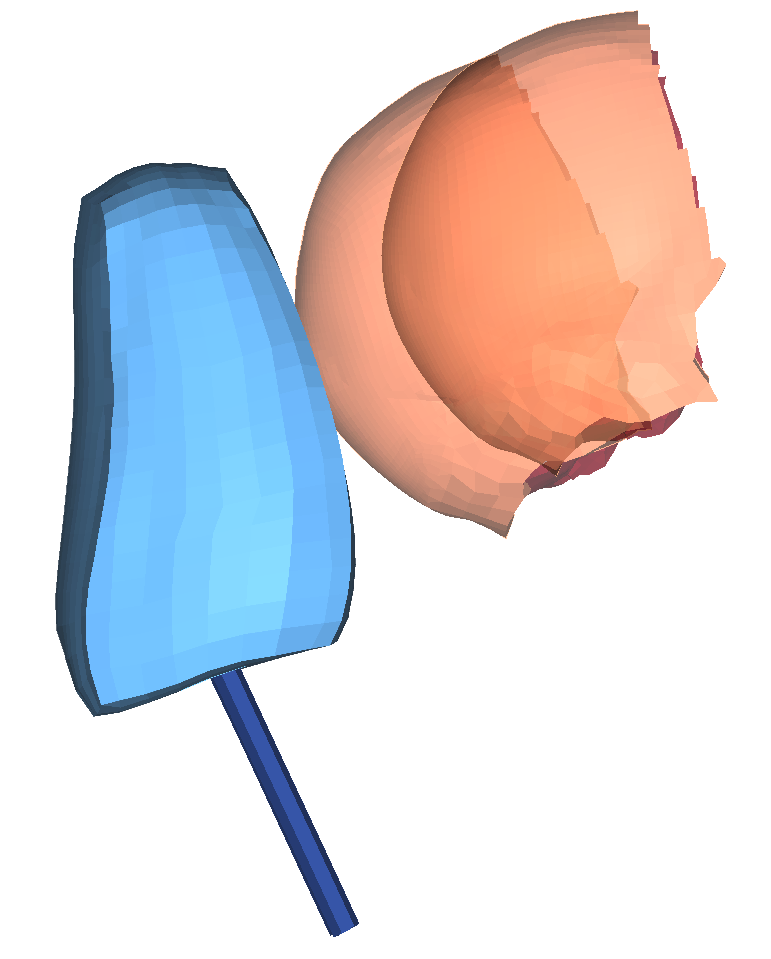};

    \node (FEmovement) at (0, 0) {\pgfuseimage{FEmovement}};

    \begin{scope}[xshift=1.5cm, yshift=-2cm]
        \draw[color=black, xshift=0.3cm, yshift=0.3cm] (-.5cm, -.5cm) rectangle (.5cm, .5cm);
        \draw[->, color=clr_1, thick] (0, 0) -- (0, 4mm) node[above] {$z$};
        \draw[->, color=clr_2, thick] (0, 0) -- (4mm, 0) node[right] {$x$};
    \end{scope}

    \begin{scope}[xshift=-1.08cm, yshift=-0.08cm, rotate=2.6]
        \draw[->, color=clr_1, thick] (0, 0) -- (0, 4mm);
        \draw[->, color=clr_2, thick] (0, 0) -- (4mm, 0);
    \end{scope}

    \begin{scope}[xshift=1.555cm, yshift=0.34cm, rotate=17]
        \draw[->, color=clr_1, thick] (0, 0) -- (0, 4mm);
        \draw[->, color=clr_2, thick] (0, 0) -- (4mm, 0);
    \end{scope}
    \node at (2.5cm, 0.34cm) {$t=0\,\textrm{s}$};

    \begin{scope}[xshift=1.245cm, yshift=0.17cm, rotate=30]
        \draw[->, color=clr_1, thick] (0, 0) -- (0, 4mm);
        \draw[->, color=clr_2, thick] (0, 0) -- (4mm, 0);
    \end{scope}
    \node at (1.5cm, -0.2cm) {$t=1\,\textrm{s}$};

    \begin{scope}
        [
            shift={(FEmovement.east)},
            xshift=1cm,
            anchor=west,
            name=plot_scope,
        ]
        \def\gap{0.3cm}
        \begin{axis}
            [
                myplot,
                at={(0, \gap)},
                ymin=-50, ymax=300,
                ytick={100, 200},
                anchor=south west,
                ylabel={Force [N]},
                name=forceplot,
                legend style={
                        at={(0.5, 1.7cm)},
                        anchor=south,
                        legend columns=3,
                    },
            ]
            \addplot[
                color=clr_2,
                thick,
            ]
            table[x=time, y=force_headrestobserver_1, col sep=comma] {img/FEmovement_forcetable.dat};
            \addlegendentry{x}
            \addlegendimage{line legend, thick, color=clr_3}
            \addlegendentry{y}
            \addplot[
                color=clr_1,
                thick,
            ]
            table[x=time, y=force_headrestobserver_3, col sep=comma] {img/FEmovement_forcetable.dat};
            \addlegendentry{z}
        \end{axis}
        \begin{axis}
            [
                myplot,
                at={(0, -\gap)},
                ymin=-5, ymax=30,
                ytick={10, 20},
                anchor=north west,
                ylabel={Moment [Nm]},
                name=momentplot,
            ]
            \addplot[
                color=clr_3,
                thick,
            ]
            table[x=time, y=moment_headrestobserver_2, col sep=comma] {img/FEmovement_momenttable.dat};
        \end{axis}
    \end{scope}
    \node[anchor=north, yshift=-.1cm] (alabel) at (FEmovement.south) {(a)};
    \node[anchor=north] at (momentplot.south |- alabel.north) {(b)};
\end{tikzpicture}
  \caption{Exemplary FE simulation of head movement toward the headrest. In (a), snapshots at times $t = 0\, \textrm{s}$ and $t = 1\, \textrm{s}$ are shown. In (b), the relevant coordinates of the interaction forces and moments acting on the head, given in the headrest coordinate system, are plotted. Initially, the head is separated from the headrest. First contact occurs at $t = 0.29\, \textrm{s}$.}
  \label{fig:FEmovement}
\end{figure}
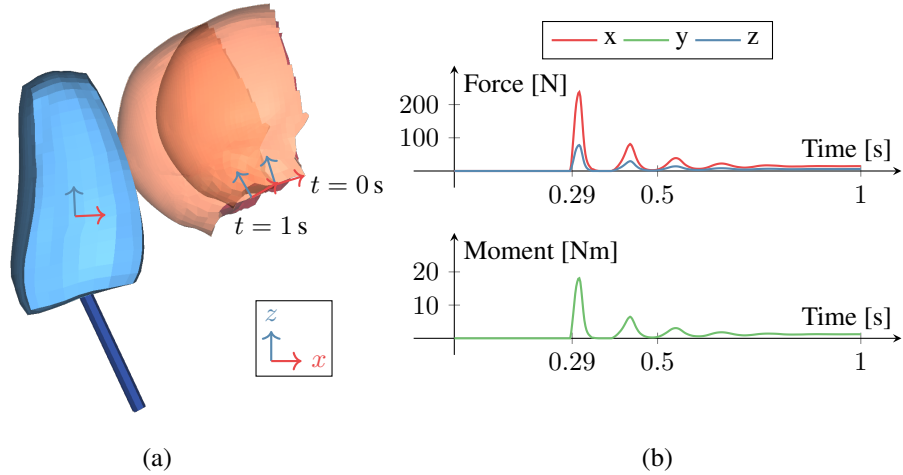

By moving the headrest bar and the head according to the MBS simulation, the interaction with the headrest cushion is replayed.
Because the bodies do not move in the second phase, the FE simulation is only performed for the first phase, i.e., up to $t=1\,\textrm{s}$.
The FE model is simulated using LS-DYNA and information about the interaction is extracted from the simulation results using a post-processing toolchain in MATLAB. 
This toolchain reads the binary output files of the FE simulation software and extracts the kinematics and the nodal contact forces for the head.
Then, the contact forces are summed to form the total contact forces and moments acting on the human body part.
With this, value pairs of the kinematics and the interaction forces and moments can be extracted for transient simulations of the occupant-seat interaction and the dataset for the surrogate model can be generated.
The nodal positions and forces are written to the binary output file with a time step size of $0.0025\,\textrm{s}$.
Therefore, every FE simulation generates a dataset of $401$ data points.
On a standard laptop, the simulation of one FE model takes approximately $3$ minutes.
Figure~\ref{fig:FEmovement} (a) shows an exemplary simulation of head movement toward the headrest.
Initially, the head is separated from the headrest and moves along a circular path toward it.
The evolution of the interaction forces and moments, depicted in Figure~\ref{fig:FEmovement} (b), indicates that first contact occurs at $t = 0.29\, \textrm{s}$.

\subsection{Multilayer Perceptron}
\begin{figure}[ht]
  \centering
  \tikzsetnextfilename{mlp}
  \begin{tikzpicture}[
    node distance=0.5cm and 0.5cm,
    layer/.style={draw, rectangle, minimum width=1.5cm, minimum height=1cm, align=center},
    hiddenlayer/.style={layer, fill=gray!10},
    arrow/.style={-{>[length=1.5mm]}},
    ]
    \node[layer] (input) {
        $\bm{r}_\textrm{rel}$\\
        $\bm{v}_\textrm{rel}$\\
        $\bm{T}^{(6)}_\textrm{rel}$\\
        $\bm{\omega}_\textrm{rel}$\\
        $\left|\bm{r}_\textrm{rel}\right|^2$\\
        $\left|\bm{v}_\textrm{rel}\right|^2$\\
        $\left|\alpha_\textrm{rel}\right|^2$\\
        $\left|\bm{\omega}_\textrm{rel}\right|^2$\\
        (19)
    };
    \node[anchor=south] (input label) at (input.north) {Input};

    \node[hiddenlayer,  right=of input, yshift=1cm] (hidden11) {\textit{Softplus}\\(6)};
    \node[anchor=south] (hidden11 label) at (hidden11.north) {Hidden};
    \node[hiddenlayer,  right=of hidden11] (hidden12) {\textit{Softplus}\\(6)};
    \node[anchor=south] (hidden12 label) at (hidden12.north) {Hidden};
    \node[hiddenlayer,  right=of hidden12] (hidden13) {\textit{Softplus}\\(4)};
    \node[anchor=south] (hidden13 label) at (hidden13.north) {Hidden};

    \node[layer,  right=of hidden13] (output1) {$\bm{F}_\textrm{rel}$\\\textit{Identity}\\(3)};
    \node[anchor=south] (output1 label) at (output1.north) {Output};

    \node[hiddenlayer,  right=of input, yshift=-1cm] (hidden21) {\textit{Softplus}\\(6)};
    \node[anchor=south] (hidden21 label) at (hidden21.north) {Hidden};
    \node[hiddenlayer,  right=of hidden21] (hidden22) {\textit{Softplus}\\(6)};
    \node[anchor=south] (hidden22 label) at (hidden22.north) {Hidden};
    \node[hiddenlayer,  right=of hidden22] (hidden23) {\textit{Softplus}\\(6)};
    \node[anchor=south] (hidden23 label) at (hidden23.north) {Hidden};
    \node[hiddenlayer,  right=of hidden23] (hidden24) {\textit{Softplus}\\(4)};
    \node[anchor=south] (hidden24 label) at (hidden24.north) {Hidden};

    \node[layer,  right=of hidden24] (output2) {$\bm{M}_\textrm{rel}$\\\textit{Identity}\\(3)};
    \node[anchor=south] (output2 label) at (output2.north) {Output};

    \foreach \i/\j in {input.east |- hidden11.west/hidden11.west, hidden11/hidden12, hidden12/hidden13, hidden13/output1}
    \draw[arrow] (\i) -- (\j);
    \foreach \i/\j in {input.east |- hidden21.west/hidden21.west, hidden21/hidden22, hidden22/hidden23, hidden23/hidden24, hidden24/output2}
    \draw[arrow] (\i) -- (\j);

\end{tikzpicture}
  \caption{Architecture of the two multilayer perceptrons used to approximate the contact forces and moments acting on the head. The activation function of each layer is written in italics and the dimensionality is given in brackets.}
  \label{fig:mlp}
\end{figure}
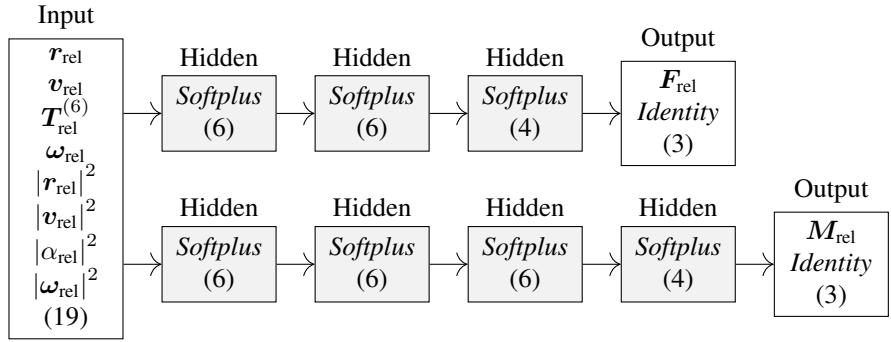

To learn the total force~$\bm{F}_\textrm{rel}$ and moment~$\bm{M}_\textrm{rel}$ acting on the body part, two individual surrogate models are trained using a multilayer perceptron (MLP) with three and four hidden layers, respectively.
The architecture of the MLPs with the dimensionality of every layer in brackets are shown in Figure~\ref{fig:mlp}.
The softplus activation function is used in the hidden layers because it is differentiable and in contrast to, e.g., the softmax or sigmoid function, it does not saturate and thus is expected to be more suitable for learning a contact situation.
The output layer uses the identity activation function to not restrict the output range of the regression model.
The input features consist of the relative position~$\bm{r}_\textrm{rel}$, the relative velocity~$\bm{v}_\textrm{rel}$, the relative orientation~$\bm{T}^{(6)}_\textrm{rel}$ represented with the first two columns of $\bm{T}_\textrm{rel}$ (see \cite{ZhouEtAl19} for an overview of rotation representations in neural networks), the relative angular velocity~$\bm{\omega}_\textrm{rel}$, and the square terms $\left|\bm{r}_\textrm{rel}\right|^2$, $\left|\bm{v}_\textrm{rel}\right|^2$, $\left|\alpha_\textrm{rel}\right|^2$, and $\left|\bm{\omega}_\textrm{rel}\right|^2$.
Hereby, $\alpha_\textrm{rel}$ is the angle between the head and headrest coordinate system, calculated with
\begin{equation}
  \alpha_\textrm{rel} = \arccos\left(\frac{\textrm{Trace}\left(\bm{T}_\textrm{rel}\right)-1}{2}\right).
\end{equation}
All of these vectors are provided to the MLP in the current headrest coordinate system and are normalized to zero mean and unit variance.
The output of the MLPs are the total force and moment acting on the body part in the headrest coordinate system.
The available simulation data is split into a set used for training and a test set with a ratio of $70\%$ to $30\%$.
Hereby, the individual simulations are not split but assigned for the training or to the test set as a whole.
This way, all data points in the test set are truly independent of the data used for training.
The data used for training is once again split into a training and a validation set with a ratio of $80\%$ to $20\%$.
Hereby, every data point is individually assigned to the training or validation set.
The validation data is used for monitoring the training process and for early stopping.
The MLPs are individually trained using the Adam optimizer with a learning rate of $0.01$, a batch size of $1024$, and early stopping with a patience of $6$ epochs.

\subsection{Active Learning Loop\label{sec:active_learning_loop}}
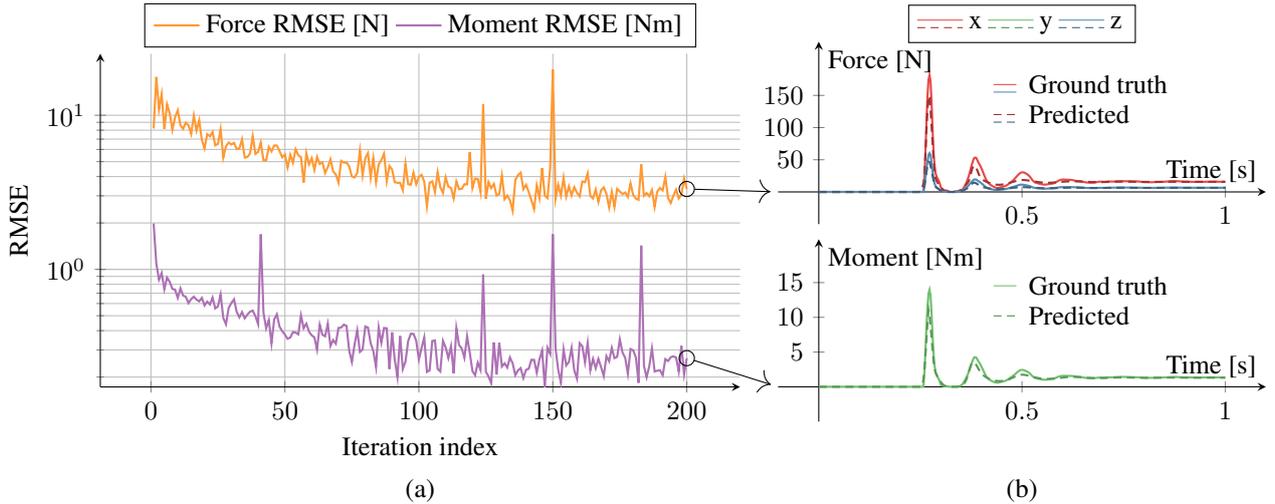
\begin{figure}
  \centering
  \tikzsetnextfilename{activelearningloop_rmse}
  \pgfplotsset{
    myplot/.style={
            xlabel={Time [s]},
            xmin=0, xmax=1,
            xtick={0,0.5, 1},
            axis lines=left,
            axis lines=middle,
            enlarge x limits,
            enlarge y limits={0.05, upper},
            width=8cm,
            height=4cm,
        },
    custom legend entry/.style={
            legend image code/.code={
                    \def\gap{0.05cm}
                    \def\length{0.6cm}
                    \draw[#1] plot coordinates {
                            (0cm,\gap)
                            (\length,\gap)
                        };
                    \draw[color=#1!70!black, densely dashed] plot coordinates {
                            (0cm,-\gap)
                            (\length,-\gap)
                        };
                }
        },
}

\begin{tikzpicture}[
    arrow/.style={-{>[length=1.5mm]}},
    ]
    \begin{axis}
        [
            2dplot,
            xlabel={Iteration index},
            ylabel={RMSE},
            enlarge x limits,
            enlarge y limits={0.05, upper},
            name=rmseplot,
            ymode=log,
            legend style={
                    at={(0.5, 4.5cm)},
                    anchor=south,
                    legend columns=2,
                },
        ]
        \addplot[
            color=clr_4,
            thick,
        ]
        table[x=index, y=forcermse, col sep=comma] {img/activelearningloop_rmse.dat};
        \addlegendentry{Force RMSE [N]};
        \addplot[
            color=clr_5,
            thick,
        ]
        table[x=index, y=momentrmse, col sep=comma] {img/activelearningloop_rmse.dat};
        \addlegendentry{Moment RMSE [Nm]};
        \coordinate (forcermselast) at (axis cs: 200, 3.322576845073355);
        \coordinate (momentrmselast) at (axis cs: 200, 0.26526395810879577);
    \end{axis}
    \begin{scope}
        [
            shift={(rmseplot.east)},
            xshift=0.5cm,
            yshift=-0.15cm,
            anchor=west,
            name=plot_scope,
        ]
        \def\gap{0.1cm}
        \begin{axis}
            [
                myplot,
                at={(0, \gap)},
                ymin=-50, ymax=220,
                ytick={50, 100, 150},
                anchor=south west,
                ylabel={Force [N]},
                name=forceplot,
                legend style={
                        at={(0.5, 2.4cm)},
                        anchor=south,
                        legend columns=3,
                    },
            ]
            \addplot[
                color=clr_2,
                thick,
                forget plot,
            ]
            table[x=time, y=forces_ground_truth_1, col sep=comma] {img/last_cycle_force_moment.dat};
            \addplot[
                color=clr_2!70!black,
                thick,
                densely dashed,
                forget plot,
            ]
            table[x=time, y=forces_predicted_1, col sep=comma] {img/last_cycle_force_moment.dat};
            \addplot[
                color=clr_1,
                thick,
                forget plot,
            ]
            table[x=time, y=forces_ground_truth_3, col sep=comma] {img/last_cycle_force_moment.dat};
            \addplot[
                color=clr_1!70!black,
                thick,
                densely dashed,
                forget plot,
            ]
            table[x=time, y=forces_predicted_3, col sep=comma] {img/last_cycle_force_moment.dat};
            \addlegendimage{custom legend entry=clr_2}
            \addlegendentry{x}
            \addlegendimage{custom legend entry=clr_3}
            \addlegendentry{y}
            \addlegendimage{custom legend entry=clr_1}
            \addlegendentry{z}
            \coordinate (forceplotwest) at (axis cs: -0.1, 0);
            \node (groundtruthlegend) at (2cm,1.8cm) {
                \def\gap{0.05cm}
                \def\length{0.3cm}
                \draw[clr_2] plot coordinates {
                        (0cm,\gap)
                        (\length,\gap)
                    };
                \draw[color=clr_1] plot coordinates {
                        (0cm,-\gap)
                        (\length,-\gap)
                    };
                \node[right, xshift=-0.4cm] at (\length, \gap) {Ground truth};
            };
            \node (predictedlegend) at (groundtruthlegend.south west) [anchor=north west, yshift=-0.15cm] {
                \def\gap{0.05cm}
                \def\length{0.3cm}
                \draw[clr_2!70!black, densely dashed] plot coordinates {
                        (0cm,\gap)
                        (\length,\gap)
                    };
                \draw[color=clr_1!70!black, densely dashed] plot coordinates {
                        (0cm,-\gap)
                        (\length,-\gap)
                    };
                \node[right, xshift=-0.4cm] at (\length, \gap) {Predicted};
            };
        \end{axis}
        \begin{axis}
            [
                myplot,
                at={(0, -\gap)},
                ymin=-5, ymax=20,
                ytick={5, 10, 15},
                anchor=north west,
                ylabel={Moment [Nm]},
                name=momentplot,
            ]
            \addplot[
                color=clr_3,
                thick,
            ]
            table[x=time, y=moments_ground_truth_2, col sep=comma] {img/last_cycle_force_moment.dat};
            \addplot[
                color=clr_3!80!black,
                thick,
                densely dashed,
            ]
            table[x=time, y=moments_predicted_2, col sep=comma] {img/last_cycle_force_moment.dat};
            \coordinate (momentplotwest) at (axis cs: -0.1, 0);
            \node (groundtruthlegend) at (2cm,1.8cm) {
                \def\gap{0cm}
                \def\length{0.3cm}
                \draw[clr_3] plot coordinates {
                        (0cm,\gap)
                        (\length,\gap)
                    };
                \node[right, xshift=-0.4cm] at (\length, \gap) {Ground truth};
            };
            \node (predictedlegend) at (groundtruthlegend.south west) [anchor=north west, yshift=-0.15cm] {
                \def\gap{0cm}
                \def\length{0.3cm}
                \draw[clr_3!70!black, densely dashed] plot coordinates {
                        (0cm,\gap)
                        (\length,\gap)
                    };
                \node[right, xshift=-0.4cm] at (\length, \gap) {Predicted};
            };
        \end{axis}
    \end{scope}

    \draw (forcermselast) circle [radius=0.1cm];
    \draw (momentrmselast) circle [radius=0.1cm];
    \draw[arrow, shorten <=0.1cm, shorten >=0.1cm] (forcermselast) -- (forceplotwest);
    \draw[arrow, shorten <=0.1cm, shorten >=0.1cm] (momentrmselast) -- (momentplotwest);

    \node[anchor=north, yshift=-1.1cm] (alabel) at (rmseplot.south) {(a)};
    \node[anchor=north] at (momentplot.south |- alabel.north) {(b)};
\end{tikzpicture}
  \caption{
    (a) Evolution of the root mean squared error (RMSE) between the predicted and the true contact forces and moments based on the head trajectory of the current iteration.
    (b) Relevant coordinates of the true (solid line) and predicted (dashed line) interaction forces and moments acting on the head for the trajectory of the last iteration.
  }
  \label{fig:activelearningloop_rmse}
\end{figure}

The active learning loop as described in Section~\ref{sec:active_learning} is implemented using a custom Python library that orchestrates the concatenation of the training data, the ML model training, the MBS simulation, and the FE simulation.
The Python library manages the data flow between the different components and allows the loop to be restarted whenever necessary.
As discussed in Section~\ref{sec:active_learning}, the loop is initially started with a preliminary dataset.
In this proof of concept, the preliminary dataset is generated using a simple contact model based on the penetration of a single point on the head (located at $\bm{r}_\textrm{sp}$ with respect to head coordinate system) with a plane representing the headrest (outward facing plane normal~$\bm{n}$).
For this, the DOF of the MBS model are sampled with a Latin hypercube sampling (LHS) algorithm to generate a set of $K$ configurations.
Then, the auxiliary contact model
\begin{align}
  f_i                     & = \left( p_i^2 \cdot k + \dot{p}_i \cdot d_0 \left(1-e^{-\frac{\left|p_i\right|}{p_\textrm{ref}}}\right)\right) \label{eq:auxiliary_contact_model} \\
  \bm{F}_{\textrm{aux},i} & = 
  \begin{cases}
    f_i \bm{n}, & \text{if } p_i > 0 \land f_i > 0 \\
    \bm{0},     & \text{otherwise}
  \end{cases}                                      \label{eq:auxiliary_contact_model_force}                                                                                    \\
  \bm{M}_{\textrm{aux},i} & = \bm{r}_\textrm{sp} \times \bm{F}_\textrm{aux} \label{eq:auxiliary_contact_model_moment}
\end{align}
with the penetration~$p_i$, the penetration velocity~$\dot{p}_i$, the stiffness~$k$, a reference penetration~$p_\textrm{ref}$, and the damping~$d_0$ is evaluated for every configuration $i \in \{1, \ldots, K\}$.
Hereby,~\eqref{eq:auxiliary_contact_model}-\eqref{eq:auxiliary_contact_model_moment} represents a viscoelastic non-sticking contact model with quadratic stiffness and a term for preventing high damping forces at small penetrations.
The model parameters are set to $k=6\,\frac{\textrm{N}}{\textrm{mm}^2}$, $d_0=0.1\,\frac{\textrm{Ns}}{\textrm{mm}^2}$, and $p_\textrm{ref}=2\,\textrm{mm}$.
The resulting contact forces and moments are then used to generate the preliminary dataset.
The active learning loop is started with this dataset and is run for $200$ iterations.
In Figure~\ref{fig:activelearningloop_rmse}~(a), the evolution of the root mean squared error (RMSE) between the predicted and the true contact forces and moments based on the head trajectory of the current iteration is shown.
In every cycle, a new head trajectory is generated, making it unseen by the ML model. 
Several spikes are visible in the RMSE, arising because neither the convergence of the OCP nor the convergence of the training of the ML model is guaranteed.
Nevertheless, there is a decreasing trend in the RMSE for both the contact forces and moments.
This trend indicates that the ML model is increasingly accurately replicating the contact as observed in the FE simulation, while the influence of the preliminary dataset diminishes.
Figure~\ref{fig:activelearningloop_rmse}~(b) illustrates the relevant coordinates of the true and predicted interaction forces and moments acting on the head for the trajectory of the final iteration.
The predicted values agree well with the true values from the FE simulation, especially when the head is near equilibrium.
However, the model tends to underestimate the forces and moments when the actual values are high.
This bias occurs because the preliminary dataset, although its influence is reduced, remains part of the training data, skewing the ML model towards predicting lower forces and moments at high penetrations.

\section{Conclusion and Further Work}
This work introduces an active learning approach to generate training data for a surrogate model of the occupant-seat interaction and simultaneously train the model.
A crucial benefit of this approach is the automation of the generation of feasible trajectories used for querying the oracle, i.e., the FE simulation.
The active learning loop is implemented to run without manual intervention and is applied to a head-headrest interaction scene.
Hereby, the loop is started with a preliminary dataset generated using a simple contact model.
It is shown that the approach is capable of generating meaningful training data for the specified task and that the ML model can replicate the contact forces and moments seen in the FE simulation with increasing accuracy.

A limitation of the current implementation is the need for a preliminary dataset to start the active learning loop.
Nevertheless, generating this preliminary dataset is less costly than generating the high-fidelity training data manually.
When the loop is started with an adjusted task definition, results from the previous loop can be used as a warm start instead of the preliminary dataset.
Additionally, the current implementation does not include any exploration strategy.
To achieve this, the provided task could be extended to include an additional term to encourage the generation of more diverse trajectories.
Further work will focus on overcoming these limitations and applying the active learning method to more complex tasks and models.

In conclusion, the proposed approach is a promising method to reduce the manual effort involved in generating training data for data-driven surrogate models that require costly transient simulations.
It represents one step toward the goal of predicting human occupant behavior using optimally controlled MBS models.

\begin{acknowledgement}
  The study was funded by Deutsche Forschungsgemeinschaft (DFG, German Research Foundation) under Fraunhofer and DFG-Trilateral Transfer Project “EMMA4Drive” grant number 440904784.
  We thank the Deutsche Forschungsgemeinschaft (DFG, German Research Foundation) for supporting this work by funding - EXC2075 - 390740016 under Germany's Excellence Strategy. We acknowledge the support by the Stuttgart Center for Simulation Science (SimTech).
\end{acknowledgement}

\providecommand{\WileyBibTextsc}{}
\let\textsc\WileyBibTextsc
\providecommand{\othercit}{}
\providecommand{\jr}[1]{#1}
\providecommand{\etal}{~et~al.}


\begin{thebibliography}{[10]}
  
  \bibitem{SunCaoTang21}%
  \textsc{X.~Sun},  \textsc{S.~Cao},  and  \textsc{P.~Tang},
  \jr{Applied Ergonomics} \textbf{90}, 103238 (2021).
  
  
  \bibitem{BubbEtAl06}%
  \textsc{H.~Bubb},  \textsc{F.~Engstler},  \textsc{F.~Fritzsche},
  \textsc{C.~Mergl},  \textsc{O.~Sabbah},  \textsc{P.~Schaefer},  and
  \textsc{I.~Zacher},
  \jr{International Journal of Human Factors Modelling and Simulation}
  \textbf{1}(1), 140--157 (2006).
  
  
  \othercit
  \bibitem{KlugEllway21}%
  \textsc{C.~Klug} and  \textsc{J.~Ellway},
  {E}uropean {N}ew {C}ar {A}ssessment {P}rogramme ({Euro NCAP}): Technical
  Bulletin 024 -- Pedestrian Human Model Certification, Version 3.0.1 ({Euro
      NCAP}, Leuven, Belgium, November 2021).
  
  
  \othercit
  \bibitem{ObentheuerEtAl23}%
  \textsc{M.~Obentheuer},  \textsc{N.~Fahse},  \textsc{M.~Harant},
  \textsc{M.~Kleer},  \textsc{F.~Kempter},  \textsc{R.~Reinhard},
  \textsc{M.~Roller},  \textsc{S.~Bj{\"o}rkenstam},  \textsc{J.~Fehr},  and
  \textsc{J.~Linn},
  {EMMA4Drive}: A digital human model for occupant simulation in dynamic driving
  maneuvers,
  in: ECCOMAS Thematic Conference on Multibody Dynamics, Lisbon, Portugal,
  (2023).
  
  
  \othercit
  \bibitem{RollerEtAl20}%
  \textsc{M.~Roller},  \textsc{V.~Dörlich},  \textsc{J.~Linn},
  \textsc{S.~Björkenstam},  and  \textsc{P.~Mårdberg},
  A digital human model for the simulation of dynamic driving maneuvers,
  in: DHM2020: Proceedings of the 6th International Digital Human Modeling,
  edited by L.~Hanson, D.~Högberg,  and E.~Brolin (2020).
  
  
  \othercit
  \bibitem{RollerEtAl17}%
  \textsc{M.~Roller},  \textsc{S.~Bj{\"o}rkenstam},  \textsc{J.~Linn},  and
  \textsc{S.~Leyendecker},
  Optimal control of a biomechanical multibody model for the dynamic simulation
  of working tasks,
  in: ECCOMAS Thematic Conference on Multibody Dynamics, Prague, Czech Republic,
  (2017).
  
  
  \bibitem{LeyendeckerEtAl10}%
  \textsc{S.~Leyendecker},  \textsc{S.~Ober-Bl{\"o}baum},  \textsc{J.\,E.
    Marsden},  and  \textsc{M.~Ortiz},
  \jr{Optimal Control Applications and Methods} \textbf{31}(6), 505--528 (2010).
  
  
  \bibitem{WachterBiegler06}%
  \textsc{A.~W{\"a}chter} and  \textsc{L.\,T. Biegler},
  \jr{Mathematical Programming} \textbf{106}, 25--57 (2006).
  
  
  \othercit
  \bibitem{PIPER24}%
  \textsc{{PIPER}},
  Piper project software framework, 2024,
  \url{http://piper-project.org/framework}, accessed May 6, 2024.
  
  
  \othercit
  \bibitem{Settles09}%
  \textsc{B.~Settles},
  Active learning literature survey,
  Computer Sciences Technical Report 1648, 2009.
  
  
  \bibitem{BuddRobinsonKainz21}%
  \textsc{S.~Budd},  \textsc{E.\,C. Robinson},  and  \textsc{B.~Kainz},
  \jr{Medical Image Analysis} \textbf{71}, 102062 (2021).
  
  
  \bibitem{IwamotoNakahiraKimpara15}%
  \textsc{M.~Iwamoto},  \textsc{Y.~Nakahira},  and  \textsc{H.~Kimpara},
  \jr{Traffic Injury Prevention} \textbf{16}(sup1), 36--48 (2015).
  
  
  \othercit
  \bibitem{MarzouguiEtAl14}%
  \textsc{D.~Marzougui},  \textsc{D.~Brown},  \textsc{H.~Park},
  \textsc{C.~Kan},  and  \textsc{K.~Opiela},
  Development \& validation of a finite element model for a mid-sized passenger
  sedan,
  in: Proceedings of the 13th International LS-DYNA Users Conference, Dearborn,
  MI, USA,  (2014),  pp.\,8--10.
  
  
  \bibitem{Hallquist06}%
  \textsc{J.\,O. Hallquist} \etal{},
  \jr{Livermore Software Technology Corporation} \textbf{3}, 25--31 (2006).
  
  
  \othercit
  \bibitem{ZhouEtAl19}%
  \textsc{Y.~Zhou},  \textsc{C.~Barnes},  \textsc{J.~Lu},  \textsc{J.~Yang},
  and  \textsc{H.~Li},
  On the continuity of rotation representations in neural networks,
  in: Proceedings of the IEEE/CVF Conference on Computer Vision and Pattern
  Recognition,  (2019),  pp.\,5745--5753.
  
  
\end{thebibliography}
\end{document}